\documentclass[letterpaper]{article} % DO NOT CHANGE THIS
\usepackage{aaai24}  % DO NOT CHANGE THIS
\usepackage{times}  % DO NOT CHANGE THIS
\usepackage{helvet}  % DO NOT CHANGE THIS
\usepackage{courier}  % DO NOT CHANGE THIS
\usepackage[hyphens]{url}  % DO NOT CHANGE THIS
\usepackage{graphicx} % DO NOT CHANGE THIS
\usepackage{natbib}  % DO NOT CHANGE THIS AND DO NOT ADD ANY OPTIONS TO IT
\usepackage{caption} % DO NOT CHANGE THIS AND DO NOT ADD ANY OPTIONS TO IT
\usepackage{algorithm}
\usepackage{algorithmic}

\usepackage{multirow}
\usepackage{array}
\usepackage{booktabs}
\usepackage{subfigure}
\usepackage{subfiles}
\usepackage{zref-totpages}

\newcolumntype{L}[1]{>{\raggedright\let\newline\\\arraybackslash\hspace{0pt}}m{#1}}
\newcolumntype{C}[1]{>{\centering\let\newline\\\arraybackslash\hspace{0pt}}m{#1}}
\newcolumntype{R}[1]{>{\raggedleft\let\newline\\\arraybackslash\hspace{0pt}}m{#1}}

\usepackage{newfloat}
\usepackage{listings}
\DeclareCaptionStyle{ruled}{labelfont=normalfont,labelsep=colon,strut=off} % DO NOT CHANGE THIS
\floatstyle{ruled}
\newfloat{listing}{tb}{lst}{}
\floatname{listing}{Listing}
\title{Characterizing Information Seeking Events in Health-Related Social Discourse}
\title{Characterizing Information Seeking Events in Health-Related Social Discourse}

\author {
    Omar Sharif\textsuperscript{\rm 1},
    Madhusudan Basak\textsuperscript{\rm1},
    Tanzia Parvin\textsuperscript{\rm 2},
    Ava Scharfstein\textsuperscript{\rm 1},
    Alphonso Bradham\textsuperscript{\rm 1},
    Jacob T. Borodovsky\textsuperscript{\rm 3, \rm 4},
    Sarah E. Lord\textsuperscript{\rm 3, \rm 4, \rm 5},
    Sarah M. Preum\textsuperscript{\rm 1, \rm 3, \rm 4}
}
\affiliations {
    \textsuperscript{\rm 1}Department of Computer Science, Dartmouth College\\
    \textsuperscript{\rm 2}Department of Computer Science and Engineering, CUET, Bangladesh\\
    \textsuperscript{\rm 3}Center for Technology and Behavioral Health, Geisel School of Medicine, Dartmouth College\\
    \textsuperscript{\rm 4}Department of Biomedical Data Science, Geisel School of Medicine, Dartmouth College\\
  \textsuperscript{\rm 5}Department of Psychiatry, Dartmouth Health\\
    
    \{omar.sharif.gr, sarah.masud.preum\}@dartmouth.edu
}

\usepackage{bibentry}
\begin{document}

\maketitle

\begin{abstract}
Social media sites have become a popular platform for individuals to seek and share health information. Despite the progress in natural language processing for social media mining, a gap remains in analyzing health-related texts on social discourse in the context of events. Event-driven analysis can offer insights into different facets of healthcare at an individual and collective level, including treatment options, misconceptions, knowledge gaps, etc. This paper presents a paradigm to characterize health-related information-seeking in social discourse through the lens of events. Events here are board categories defined with domain experts that capture the trajectory of the treatment/medication. To illustrate the value of this approach, we analyze Reddit posts regarding medications for \textit{\textsc{Opioid Use Disorder}} (OUD), a critical global health concern. To the best of our knowledge, this is the first attempt to define event categories for characterizing information-seeking in OUD social discourse. Guided by domain experts, we develop \textit{TREAT-ISE}, a novel multilabel treatment information-seeking event dataset to analyze online discourse on an event-based framework. This dataset contains Reddit posts on information-seeking events related to recovery from OUD, where each post is annotated based on the type of events. We also establish a strong performance benchmark (77.4\% F1 score) for the task by employing several machine learning and deep learning classifiers. Finally, we thoroughly investigate the performance and errors of ChatGPT on this task, providing valuable insights into the LLM's capabilities and ongoing characterization efforts.
\end{abstract}
%Subsequently, we perform a qualitative investigation to surface several potential clinical insights.

\section{Introduction}
% \begin{itemize}
%     \item Discuss about the importance of data \cite{rogers-2021-changing}
%     \item Convince why data from such a niche domain is important and what are the social impact
%     \item how our data is adding values. 
% \end{itemize}

About 70\% of people in the USA rely on social media to connect with peers and share information to navigate aspects of their lives \cite{kanchan2023social}, such as finance, professional development, and health. Approximately 72\% of adult internet users engage in online searches to find information about various health issues, either for themselves or on behalf of others. Individuals from diverse socio-demographic backgrounds and medical conditions often resort to these platforms for sharing and seeking information during different phases of their healthcare journey \cite{Neely2021-ji}. As a result, online platforms offer a unique opportunity to surface and contextualize information needs, knowledge gaps, treatment perceptions, conflicting information, and misconceptions as illustrated in prior works \cite{EDOOSAGIE2020103770,chen2021social,gatto2023scope, acm_review}.
%self-reported   by a large number of individuals with diverse treatment trajectories
%Dou_Barman-Adhikari_Fang_Yadav_2021
% The anonymity cloak offered by these platforms (e.g., Reddit) helps people to seek or share information, give suggestions, and express opinions honestly. Gathering such information on a large scale through surveys or other means is challenging. Research indicates that quantitative and qualitative analysis of these data presents a novel way to comprehend different stigmatized topics such as mental health, depression, and substance use disorder \cite{romano2023themedriven}. 
%Dou_Barman-Adhikari_Fang_Yadav_2021

%The tendency to seek and share information on anonymous platforms (e.g., Reddit) is comparatively higher for stigmatized topics such as mental health and substance use disorder. This might be because might people find anonymous social platforms appropriate for seeking information, giving suggestions, and sharing experiences without revealing their identity. 

% In the United States, substance use disorder (e.g., opioid abuse) has become one of the biggest societal problems. 

Event analysis is a well-studied NLP problem that can be leveraged to get insights into online health discourse \cite{ma-etal-2023-dice, naik-etal-2017-extracting}. Analyzing posts from social support groups through the lens of events can give us a nuanced understanding of health information needs. For example, consider the following post seeking information on post-operative pain relief. 

% \begin{figure}[h!]
%   \centering
%   \includegraphics[width =1\linewidth]{Figures/event-extract.png}
%  \caption{Sample information-seeking event extraction from Reddit post. }
%  \label{exmaple}
% \end{figure}

\begin{quote}
   I just had knee surgery about a week ago, and my \textbf{pain meds }(\textbf{Gabapentin}) are not cutting it! I am having \textbf{sleep troubles} and getting pretty \textbf{anxious} about recovery. I am considering \textbf{Kratom}. Any suggestions about \textbf{how much I can take per day} and \textbf{how often}?
\end{quote}
% \begin{quote}
%    How can I manage my \underline{back-pain}? I \textbf{take} \underline{1-2 mg} \underline{naproxen} \underline{at night}. However, I am having \underline{sleep troubles}. If I take it in the morning, then I feel \underline{groggy}! I had \underline{surgery} on my back five months ago. %What should I do now? 
% \end{quote}
This post mentions multiple key events, e.g., taking medication (\textit{Gabapentin}), and experiencing psychophysical effects (\textit{pain} or \textit{anxiety}). Such event analysis based on a large number of samples can reveal insights into different aspects of treatment at both individual and collective levels (e.g., \textit{how many people report a new or rare side effect during pain management}), the perceived value of treatment (e.g., \textit{ineffective pain medication}),  self-treatment strategies (e.g., \textit{self-dosing Kratom}), knowledge gaps and concern (\textit{rare or new side effects of treatment}), and misconceptions (e.g., \textit{self-dosing Kratom is safe for pain relief}). However, a significant gap still exists in performing event-driven analysis on health discourse in online communities, including social media. 

%Suppose the possible event arguments for the trigger word ``take'' are `\textit{medication}: suboxone', `\textit{dosage}: 1-2 mg', `\textit{timing}: at night', `\textit{frequency}: per day', and `\textit{side effects}: sleep troubles'. 

% To demonstrate the value of event-driven analysis, we investigate online discourse on recovery from opioid use disorder (OUD). OUD continues to be the leading cause of death in the US, creating a huge socio-economic burden. Studies show that managing the opioid crisis costs a staggering 1.02 trillion dollars annually \cite{FLORENCE2021108350}. However, stigma towards substance dependency and recovery, limited access to care, and distrust of the traditional healthcare system make people increasingly resort to social groups to seek recovery support \cite{10.1001/jama.2014.2147}. Medications for opioid use disorder (MOUD) are the most effective treatment option for OUD and can save lives \cite{mooney2020patient}. Thus pseudo-anonymous social media platforms like Reddit present a unique opportunity to gain insights into MOUD-based recovery. We chose Reddit as it has significant reach to the US population and emphasizes anonymity, making it one of the largest platforms to capture rich content on stigmatized topics such as mental health, sexual abuse disclosure, and substance use disorder \cite{romano2023themedriven, 10.1145/3359249,sharma2020engagement}. 

To showcase the significance of event-driven analysis, we explore online discussions regarding recovery from opioid use disorder (OUD), a critical concern with substantial societal impact. OUD remains a leading cause of mortality in the US, incurring a massive economic toll, estimated at 1.02 trillion dollars annually \cite{FLORENCE2021108350}. Existing challenges, including stigma around addiction, limited healthcare access, and distrust of traditional systems, drive individuals towards seeking recovery support within social groups \cite{10.1001/jama.2014.2147}. Medications for opioid use disorder (MOUD) offer a vital treatment avenue, capable of saving lives \cite{mooney2020patient}. Pseudo-anonymous platforms like Reddit, known for its wide US user base and emphasis on anonymity, provide a unique lens into MOUD-based recovery insights. Reddit's reach and focus on sensitive topics, such as mental health and substance use disorder, position it as a significant source for rich content \cite{romano2023themedriven, 10.1145/3359249,sharma2020engagement}.

Event analysis on social media text confronts distinct hurdles. Defining and standardizing event types pose a challenge, especially considering the influence of domain-specific factors that determine event relevance—\textit{clinical} events differing from \textit{career} events, for instance. Moreover, equivalent events might be expressed in vastly dissimilar colloquial terms. Remarkably, there exists no dataset for delving into event analysis within online discourses.

% However, given the rich volume and variety of online discourse, performing event analysis on social media texts has its unique challenge, such as defining and normalizing event types. Particularly, domain-specific aspects significantly impact which events are relevant (e.g., events for \textit{parenting} might differ from those for \textit{mental health}). Also, similar events can be narrated using drastically different colloquial language.  To the best of our knowledge, no dataset exists to explore event analysis from online discourse. 
% events are often shaped by 
% , colloquial language for different domainsnoisy data, and lack of relevant labeled datasets \cite{li-etal-2014-major}.

This paper introduces an event-based framework for online discourse analysis. We study Reddit posts on MOUD, a critical topic amid the opioid crisis. Collaborating with experts, we define information-seeking events, create annotation guidelines, and curate a unique labeled dataset. Using the event schema and info-seeking posts, we explore information quality systematically. Comprehensive insight into treatment needs relies on classifying data into specific events. We frame identifying core events in posts as a multi-label, multi-class classification challenge. We assess the dataset with advanced text classifiers including large language models.
Our major contributions are as follows.
\begin{itemize}
   
    % \item \textbf{Design:} In collaboration with domain experts, we propose a treatment information-seeking event (ISE) schema that can help to understand the OUD treatment trajectory. 
    \item \textbf{Resource:} Based on guidance from domain experts, we propose a treatment information-seeking event (ISE) schema that can help to understand the OUD treatment trajectory. Leveraging this schema, we develop \textit{TREAT-ISE}, a multilabel dataset comprising human-annotated samples. The novel dataset, annotation guidelines, and associated code will accelerate further research in online health discourse analysis\footnote{All the resources are available at \url{https://github.com/omar-sharif03/AAAI-2024}}. 

     \item \textbf{Social:} We focus on a highly vulnerable population, i.e., individuals considering or undergoing OUD recovery, which has received little attention in previous work, and characterize their self-reported MOUD treatment information needs. The dataset and other outcomes can complement traditional electronic health records and survey data and capture the real-world complexity of recovery. 
    \item \textbf{Benchmarking:} We investigate the performance of ten off-the-shelf machine learning and deep learning models for this task. Furthermore, we thoroughly assess the effectiveness of ChatGPT, thereby uncovering the potential scope of ChatGPT and state-of-the-art text classifiers for such complex, knowledge-intensive discourse analysis. 
\end{itemize}

\section{Related Work}
\subsection{Social Media and Substance Use Disorder}
Social media platforms offer individuals opportunities to share the different events of their lived experiences, such as addiction, logistical barriers, treatment strategy, the experience of psychophysical effects, and more \cite{info:doi/10.2196/43349}. Prior studies have utilized these data to understand different types of substance usage, including cannabis, alcohol, opioids, and others \cite{Lavertu2021.04.01.21254815, CHEN2022100061}. \citeauthor{OpioidRecovery} \shortcite{OpioidRecovery} tried to uncover alternative treatment options for OUD by analyzing the opioid discourse on Reddit. \citeauthor{romano2023themedriven} \shortcite{romano2023themedriven} presented a framework for extracting keywords and applied it to extract insights about OUD recovery from Reddit. Another related study by   \citeauthor{Balsamo_Bajardi_De} \shortcite{Balsamo_Bajardi_De} investigated how much online social community can support individuals undergoing opioid usage. Our work differs from existing studies both in task formulation and scope. We focus on events as the primary analytical lens, encompassing diverse categories to study OUD discourse.

\subsection{Event Analysis for Health-related Discourse}
%Event analysis is an information extraction task that helps to understand unstructured texts.
Event analysis includes two main subtasks: \textit{event detection:} identifying trigger terms and event type, and \textit{argument extraction:} extracting event arguments from texts and assigning roles to arguments based on event type \cite{ma-etal-2023-dice}. %Most event analysis tasks are formulated as token-level classification tasks with event detection and event argument extraction pipelines \cite{wadden-etal-2019-entity}. More 
 Recent approaches have adopted a text-generation paradigm, leveraging large language models to prompt the extraction of event types, triggers, and arguments \cite{li-etal-2021-document,lu-etal-2021-text2event}. Few works attempted to detect events without extracting triggers \cite{liu-etal-2019-event}. However, these models often exhibit suboptimal performance when dealing with event analysis tasks heavily reliant on domain-specific knowledge \cite{li2023evaluating}. The limited availability of training data and the complexity of domain-specific terms contribute to this issue. 

Research on event analysis from social media discourse regarding health is limited. \citeauthor{naik-etal-2017-extracting} \shortcite{naik-etal-2017-extracting} attempted to develop a disease progression timeline by analyzing patient-authored texts in social support groups. They explored the connections between medical events and users' engagement within these groups. In a similar work, \citeauthor{wen_12} \shortcite{wen_12} investigated the behavioral trajectory of participants by analyzing cancer-related events in online medical discourse. In subsequent work, \citeauthor{wen-etal-2013-extracting} \shortcite{wen-etal-2013-extracting} created a temporal tagger to extract cancer-related event dates to explore treatment trajectories. 
% \citeauthor{li-etal-2014-major} \shortcite{li-etal-2014-major} focused on detecting major life events from Twitter using congratulatory and condolence speeches. 

This work focuses on event detection rather than trigger or argument extraction since it is one of the first attempts to understand OUD discourse in the context of events. Here, we formulate the event detection task as a multi-label, multi-class classification problem \cite{BOGATINOVSKI2022117215}. We curate a dataset of information-seeking events by actively involving domain experts in the process. Moreover, we analyze ChatGPT's performance and errors, presenting valuable insights into the model's capabilities and contributing to ongoing efforts to understand its characteristics.

%\textcolor{red}{[I will try to add a short paragraph, how our work is different]}

%\section{Treatment Information-Seeking Event Schema Design }
\section{Defining Information-Seeking Events}
Our overarching goal is to characterize information-seeking events (ISE) from online discourse. These events are self-reported by individuals considering or undergoing medications for OUD (MOUD) treatment. MOUD includes Buprenorphine (e.g., Suboxone, Subutex, Sublocade), Methadone, and Naltrexone \cite{dickson2022you}. We collaborate with five domain experts to define the events that best characterize the treatment information-seeking events from different stages of the treatment journey. Our collaborators are well-versed and internationally acclaimed scholars in substance use disorder, spanning various areas such as epidemiology, public health policy, mental health, addiction psychiatry, addiction medicine, and biomedical data science. They review the collected samples and offer valuable insights into various facets of opioid recovery. Based on the guidance from the domain experts, we identify five coarse categories of events for treatment information needs. These ISE categories are: \textit{Accessing MOUD}, \textit{Taking MOUD}, \textit{Experiencing Psychophysical Effects}, \textit{Relapse or co-occurring substance usage}, and \textit{Tapering MOUD}. All of these events are prevalent in recovery using MOUD. 
% Additionally, we define \textit{Others}, category to encompass samples that seek treatment information but do not fall within the scope of the five target categories. The definitions of these categories in the context of MOUD are described below.

\begin{itemize}
    \item \textbf{Accessing MOUD (AM):} Information-seeking events related to accessing MOUD, such as concerns about insurance, pharmacy, providers, etc. Analyzing samples from this event can help to determine the common barriers people encounter during recovery using MOUD that affect treatment induction, adherence and retention. 
    
    %Logistical issues or barriers that prevent individuals from accessing MOUD, such as insurance, pharmacy, provider-specific issues, family problems, not knowing what is needed to access, not knowing where to access, etc.

      %% MOUD Administration
     \item \textbf{Taking MOUD (TM):} Information-seeking events related to MOUD regimen details, e.g., questions about timing, dosage, frequency of taking a MOUD, concerns about splitting and missing a dose.  This class can surface potential misconceptions and concerns about MOUD administration that negatively impact treatment adherence. 

    \item \textbf{Experiencing Psychophysical Effects during Recovery (EP):} Information-seeking events related to concern about potential physical and/or psychological effects during recovery. This event class covers both experienced and anticipated psychophysical effects. It can surface rare and new adverse effects of MOUD as well as prevalent psychophysical effects of MOUD, their severity and potential impact of treatment adherence. 

    \item \textbf{Relapse (RL) or co-occurring substance usage:} This class includes events that talk about relapsing or using other substances during recovery. Such substance use can be attributed to recreational purposes or for self-medication (e.g., marijuana for sleep). We follow NIDA's\footnote{https://tinyurl.com/4ckwz453} list of commonly used drugs to identify what counts as a substance. Samples of this event class can help unearth specific information individuals seek concerning recreational and medical usage of substances. 

    \item \textbf{Tapering MOUD (TP):} Information-seeking events related to reducing the dose or frequency of MOUD and eventually quitting MOUD. Although the current standard of care recommends consulting healthcare providers for tapering MOUD, individuals often resort to self-managed tapering strategies.  Analyzing events from this class can inform addiction researchers and clinicians about the context of self-managed tapering strategies (e.g., why and when people self-taper) and their effectiveness (what works for whom).
    
    \item \textbf{Others (Oth):} Information-seeking events related to other issues. 
\end{itemize}

In this work, we focus on information-seeking events from posts on social media. It should be noted that we can also analyze relevant information-providing or sharing events (i.e., comments or replies to the original post) through this lens of events. This will help us measure the availability and quality of shared information in online discourse more systematically, e.g., self-management strategies for tapering, high-dose of MOUD suggested by peers or common misinformation regarding relapse during recovery using MOUD.
%These information-seeking events collectively help to characterize the trajectory of OUD recovery.
Such systematic analysis can potentially uncover actionable insights to improve treatment adherence and outcomes.
% increase treatment induction, adherence, and retention for recovery from OUD.

%It is crucial to note that achieving a comprehensive understanding of treatment information involves two fundamental phases: (i) classifying information into specific ISEs and (ii) conducting an extensive qualitative analysis of the categorized samples. This work primarily concentrates on the initial phase, which involves effectively organizing samples into predefined ISE classes. The success of the subsequent qualitative analysis greatly relies on the accuracy and effectiveness of this sample categorization process. 

% \begin{itemize}
%     \item Introduce the task and different labels.
%     \item Definition of different labels with examples. 
%     \item Arguments on why we choose these class labels and how it is important. Make sure to point out that we have discussed with domain expert to choose these class labels. 
% \end{itemize}

\begin{table*}[ht!]
\centering
\footnotesize
\begin{tabular}{L{3.5cm} L{10.5cm} C{2.8cm} }
\hline
\textbf{Title}&\textbf{Post}&\textbf{Events}\\
\hline                   
Looking for suboxone guidance? & I take 1-2mg subs per day which is a decrease from the original dose of 8mg. Just looking for a plan of action in which to stick with to eventually get off completely. & Taking MOUD (TM), Tapering (TP) \\
\midrule
 Which Kratom strain helps with Bupe withdrawal & When I run out of my Suboxone prematurely, I like to keep Kratom on hand for my extremely low energy and excessive yawning. & Relapse (RL), Psychophysical effects (EP) \\
\hline

\end{tabular}
\caption{Sample data excerpts with titles, posts, and labels (shortened and paraphrased as per IRB guideline).}
\label{samples}
\end{table*}

\section{Dataset Collection and Annotation Strategies}
\subsection{Data collection}
We chose Reddit as the data source due to its anonymity policy and rich content on MOUD \cite{alternative_chi}. We selected r/Suboxone as our primary data source as (i) it has both the highest number of members and the number of peer interactions (e.g., number of posts, comments) among the subreddits specific to different options of MOUD; and (ii) it is strictly moderated where any irrelevant posts are removed by moderators (e.g., drug soliciting posts). So, this subreddit offers a unique chance to understand users' information needs related to a MOUD authentically. We scraped all the posts between January 2018 (as minimal interaction was observed in this subreddit prior to 2018) and August 2022 (study start time). We collected a total of 25,044 posts using the PRAW and PushShift APIs \cite{baumgartner2020pushshift}.  The collected data includes titles, posts, comments, likes, upvotes, and unique post IDs while strictly adhering to ethical considerations by not collecting/storing information that violates ethical concerns. After removing irrelevant posts (e.g., polls, link-only posts), we ended up with a corpus of 15,253 relevant posts. Among these posts, we annotate 5083 randomly selected posts.

\subsection{Data Annotation}

\textbf{Feasibility of Crowd-sourced Annotation:} Annotating the type of treatment information-seeking events is a challenging task that demands significant effort. Initially, we employed the widely-used approach of annotating through crowd-workers on Amazon Mechanical Turk \cite{mirzaei-etal-2023-real}. We selected a pool of Master qualified workers (mTurkers with high approval ratings), provided them with explicit annotation guidelines, and conducted a trial run on 300 samples. However, we encountered poor annotation quality and low inter-annotator agreement (only 40.5\%). This is because this annotation task requires a good understanding of domain knowledge and annotators need interactive, progressive training sessions to ensure they understand the nuances of different types of events. Our trial run indicates a lack of suitability of crowd-workers for such a challenging inference task. Therefore, we decided to perform in-house annotation with students and experts.

\noindent
\textbf{Annotation process:} To complete the annotation, we form a diverse group of 9 annotators: 3 undergraduates and 6 graduate students. Initially, we provided them with background knowledge on MOUD and suboxone through multiple sessions led by experts. %We trained them with definitions of TIN classes along with annotation guidelines. 
To achieve quality annotation, our primary focus was to confirm that the annotators understand what are the ISEs for OUD recovery and how a user can seek information for multiple event types in a post (details added in the appendix\footnote{https://tinyurl.com/yyrn7ywn}). Each annotator was trained for four weeks through trial annotation tasks before they started actual annotation to ensure annotators were well-versed with ISE classes and eliminate the uncertainties about annotation guidelines.  Each sample was reviewed by at least two different annotators for the annotation. Table \ref{samples} demonstrates sample posts with associated labels. 

\noindent
\textbf{Inter-annotator Agreement:} We compute the inter-annotator agreement in terms of Cohen's $\kappa$-score \cite{cohen1960coefficient}. Table \ref{kappa-score} shows the $\kappa$-score for each class where the AM class achieves the highest agreement score of 0.86, and the \textit{other} class gets the lowest (0.68). The mean $\kappa$-score of 0.76 indicates substantial agreement between the annotators. The presence of domain-specific drug names, lengthy text samples with shorthand, slang, and misspellings posed challenges during annotation. Table \ref{kappa-score} exhibits the initial agreement score between two annotators. 
It is important to mention that a domain expert reviewed all the samples after labeling by two annotators to ensure the data quality. Subsequently, resolved any confusion or annotation disagreement and rectified the labels. This domain expert is a study team member who is disjointed from the set of recruited annotators. This complies with the recommended best practice for qualitative health data annotation/coding \cite{busetto2020use}. Thus we develop \textbf{TREAT-ISE} a MOUD treatment information-seeking event dataset comprised of 5083 multilabel samples. 
\begin{table}[h!]
\centering
\footnotesize
\begin{tabular}{l|cccccc|c}
&\textbf{AM}&\textbf{TM}&\textbf{TP}&\textbf{EP} & \textbf{RL} & \textbf{Oth}&\textbf{Mean}\\
\midrule        
$\kappa$-score  & 0.86 &0.72 & 0.82 & 0.74 & 0.75 &0.68 & 0.76 \\
\hline

\end{tabular}
\caption{Classwise $\kappa$-score of TREAT-ISE.}
\label{kappa-score}
\end{table}

\begin{table}[h!]
\centering
\footnotesize
\begin{tabular}{l|ccC{1.5cm}C{1.9cm}}
\textbf{Class}&\textbf{\#Samples}&\textbf{\#Words}&\textbf{\#Unique words}&\textbf{\#Avg. words/sample}\\
\midrule        
AM & 873 &108k & 7665 & 124.16\\
TM & 1637 & 199k & 10477 &  122.07\\
TP & 1424 & 215k & 11087 & 151.40\\
EP & 1837 & 271k & 13395 & 147.75\\
RL & 1420 & 202k & 10776 & 142.33\\
Oth & 473 & 48k & 6159 & 102.62\\
\hline

\end{tabular}
\caption{Summary of different classes of the TREAT-ISE dataset. 
}
\label{statistics}
\end{table}

Table \ref{statistics} presents the statistics and lexical summary of TREAT-ISE. The dataset is imbalanced, with EP having the highest number of samples. Among the classes, EP stands out with the most words ($\approx$271k) and unique words ($\approx$ 13k), while the AM class has the lowest counts ($\approx$108k, $\approx$7.6k). TREAT-ISE stands apart from other domain-specific datasets by presenting a unique multilabel classification challenge with significantly longer average sample lengths (ranging from $122$ to $151$). The average sample length for similar multilabel classification tasks is less than 50 \cite{Patwa_2021,van-aken-etal-2018-challenges}. In the ablation studies, we present a few insights into how large language models handle these domain-specific long texts. %\textcolor{blue}{Compare with a few other datasets to justify the claim that our samples are longer.}
%Five frequent words of each class with their tf-idf score are shown in table \ref{frequent-words}.
%We can see words like \textit{insurance, pharmacy, appointment} are frequent in the Access Logistics class. We also notice the presence of domain-specific drug names such as \textit{fentanyl oxycodone, kratom}. 

% \begin{table}[h!]
% \centering
% \footnotesize
% \begin{tabular}{l|C{6.8cm}}
% \textbf{Class}&\textbf{Frequent words}\\
% \midrule        
% AM & suboxone (30.32), doctor (24.37), insurance (20.53), pharmacy (20.44), appointment (17.54) \\
% TM & dose (54.33), days (39.99), subutex (28.84), 8mg (25.94), withdrawal (24.90) \\

% TP & subs (47.37), taper (44.33), time (33.60), help (30.36), jump (23.01)  \\

% EP & withdrawal (34.67), pain (30.24), sleep (27.69), life (22.29), sick (21.37)  \\

% RL & kratom (55.78), fentanyl (27.25), oxycodone (24.39), subutex (19.94), advice (17.80) \\

% OTH & test (12.30), drug (10.30), addiction (5.54), bupe (5.38), recovery (5.06) \\
% \hline
% \end{tabular}
% \caption{\label{frequent-words} Five frequent words in each class based on tf-idf score (provided within parenthesis).}
% \end{table}

% \begin{itemize}
%     \item provide different statistics of the dataset: number of samples/words/unique works in each class.
%     \item Do some analysis from different angles: gender, age, drug, dosage, duration
% \end{itemize}

\begin{table*}[h!]
\centering
\renewcommand*{\arraystretch}{1}
\footnotesize
\begin{tabular}{cccc|ccc|ccc|ccc|ccc|c}

 & \multicolumn{3}{c}{\textbf{AM}}& \multicolumn{3}{c}{\textbf{TM}}& \multicolumn{3}{c}{\textbf{TP}} & \multicolumn{3}{c}{\textbf{EP}} & \multicolumn{3}{c}{\textbf{RL}} & \\
\hline
\textbf{Model}&\textbf{P}&\textbf{R}&\textbf{F1}&\textbf{P}&\textbf{R}&\textbf{F1}&\textbf{P}&\textbf{R}&\textbf{F1}&\textbf{P}&\textbf{R}&\textbf{F1}&\textbf{P}&\textbf{R}&\textbf{F1}&\textbf{WF1}\\
\hline
\multicolumn{17}{c}{\textit{Non-transformer Baselines}}\\
\hline
LR &  0.63 & 0.68 & 0.65  & 0.61 & 0.64 & 0.63  &  0.74 &0.64 & 0.68  & 0.49 & 0.66 & 0.56  &  0.71 & 0.57 &0.63 & 0.593\\
NBSVM & 0.73 & 0.57 & 0.64 &  0.58 & 0.73 & 0.65  & 0.72  & 0.78 & 0.75  & 0.44 & 0.75 & 0.56 &  0.61 & 0.62  &    0.62 & 0.602 \\
FastText & 0.64 & 0.73 & 0.68 &  0.61 & 0.70 & 0.65  & 0.69 & 0.84 & 0.75   & 0.50 & 0.78 & 0.61 & 0.63  &    0.64 & 0.63 & 0.624 \\
BiGRU & 0.72 &0.69 & 0.70 &  0.73&0.68 & 0.70  & 0.80      & 0.84 & 0.82 & 0.61 & 0.61 & 0.61 & 0.80 & 0.72  &    0.76 & 0.702\\
\hline
\multicolumn{17}{c}{\textit{Transformer Baselines}}\\
\hline
BERT & 0.85  & 0.74  & 0.79 & \textbf{0.83} & 0.64  & 0.72 &  0.84 & 0.84 & 0.84 & \textbf{0.69} & 0.59 & 0.64 & 0.88 & 0.76 & 0.81 & 0.733  \\

%ALBERT & 0.85 & 0.78 & 0.81 & 0.87 & 0.61 & 0.71 & 0.86 & 0.84 & 0.85  & 0.72 & 0.46 & 0.56 & 0.89 & 0.76 & 0.82 & 0.725   \\

RoBERTa & 0.82 & 0.82 & 0.82 & 0.75  & 0.80  & \textbf{0.77} & 0.80 & 0.89 & 0.84 & 0.63 & 0.74  &  \textbf{0.68} & 0.88 & 0.75  & 0.81 & 0.757    \\

Distil-BERT & 0.80  & 0.69 & 0.74 & 0.81 & 0.62 & 0.70 & 0.84 & 0.78 & 0.81 & 0.68 & 0.57  &    0.62 & 0.84 & 0.76  & 0.80 & 0.711  \\

ELECTRA &  0.77 & 0.79 & 0.78 &0.80  & 0.67  & 0.73 & \textbf{0.87} & 0.84  & 0.85 &  0.65 & 0.64 &     0.65 & 0.80  & \textbf{0.88}  & 0.84 & 0.748   \\

XLNet & 0.84 & \textbf{0.82} & \textbf{0.83} & 0.79 & 0.72 & 0.75 & 0.85 & 0.84 & \textbf{0.85} & 0.59 & 0.78 & 0.67 &\textbf{0.88} & 0.85 & \textbf{0.86} & \textbf{0.774}   \\
MPNet & 0.79 & 0.81 & 0.80 & 0.80 & 0.71 & 0.75 & 0.81 & 0.85 & 0.83 & 0.68 & 0.66 & 0.67 & 0.78 & 0.82 & 0.80 & 0.751\\
\hline
\multicolumn{17}{c}{\textit{ChatGPT Baselines}}\\
\hline
ChatGPT (ZS-S) & \textbf{1.0} & 0.26  & 0.41  &  0.74 & 0.30& 0.43  &  0.67& 0.42 & 0.52  &  0.62 & 0.10 & 0.18  & 0.62 & 0.81 &  0.70  & 0.433\\
ChatGPT (ZS-L) & 0.78 &  0.61 &  0.69  &  0.68 & 0.63 & 0.65  &  0.69 & 0.67 & 0.68  &  0.70 & 0.29 & 0.41  & 0.77 & 0.53 & 0.63  & 0.581 \\
ChatGPT (FS-S) &  0.48 & 0.79 & 0.60  & 0.47 & \textbf{0.92} & 0.62  &  0.45 & \textbf{0.96} & 0.61  &  0.44 & 0.83& 0.57  & 0.65 & 0.69 &     0.67  & 0.609 \\
ChatGPT (FS-L) &  0.51  & 0.78 & 0.62  & 0.52  & 0.87 &   0.65 &  0.50 & 0.86  &  0.63  & 0.40  &   \textbf{0.92}   &   0.56  &0.66    &  0.72  &    0.69  & 0.620 \\

ChatGPT (CoT) & 0.62 & 0.78 & 0.69 & 0.49 & 0.87 & 0.62 &  0.55 &  0.89 & 0.68  &  0.49 & 0.76 & 0.60 & 0.74 & 0.56 & 0.64 & 0.631 \\
\hline            
\end{tabular}
\caption{Classwise performance for treatment information seeking event detection. WF1 indicates the weighted F1 score based on all six classes. The shorthand indicates ZS-S, ZS-L: Zero-shot (Short, Long), FS-S, FS-L: Few-shot (Short, Long), and CoT: Chain-of-Thought prompting. Due to space constraints, the models' performance in \textit{other} class is not included. }
\label{all-results}
\end{table*}
\section{Methodology}

We present a comprehensive benchmark evaluation of the TREAT-ISE dataset encompassing various methodologies, including off-the-shelf non-transformer, Transformer-based, and Large language models such as ChatGPT. These methods represent standard approaches for multilabel classification and provide a diverse range of technical implementations for thoroughness. The details of each method are described in the subsequent paragraphs. 
\begin{itemize}
    \item \textbf{Non-transformer models:} In the baseline evaluation, we explore the performance with two machine learning models: Logistic Regression (LR) \cite{tibshirani-manning-2014-robust} and Naive Bayes with Support Vector Machine (NBSVM) \cite{wang-manning-2012-baselines}. For deep learning approaches, we investigate two variants: one utilizes pretrained FastText \cite{joulin-etal-2017-bag} embeddings with a feedforward network, and the other employs a Bidirectional Gated Recurrent Unit (BiGRU). In BiGRU, embedding features are propagated to a GRU layer with 80 hidden units. The output from the last hidden layer is passed to global average pooling and max-pooling layers. Subsequently, the outputs of the pooling layers are concatenated and passed for classification. 
     %In LR, we use `lbfgs' optimizer with `l2' regularizer while NBSVM combines the power of NB and SVM with interpolation value $\beta=0.25$. Embedding features are utilized to train these models. 
    \item \textbf{Transfomer-based models:} In recent years, transfomer-based \cite{NIPS2017_3f5ee243} models have achieved state-of-the-art performance on various NLP tasks. We employ six transformer-based pre-trained models to benchmark the multilabel ISE classification task. These include Bidirectional Encoder Representations from Transformers (BERT) \cite{devlin-etal-2019-bert}, 
    %a Lite version of BERT with lower memory consumption (ALBERT) \cite{lan2020albert}, 
    a distilled variant of BERT (DistilBERT) \cite{sanh2020distilbert}, robust BERT architectures with more training data RoBERTa \cite{liu2019roberta}, ELECTRA \cite{clark2020electra}, a model with generalized autoregressive pertaining (XLNet) \cite{yang2020xlnet} and MPNet \cite{song2020mpnet}. All the models are sourced from the Huggingface library. Subsequently, fine-tuned on our dataset for 10 epochs with a learning rate $2e^{-5}$ and batch size 16. The intermediate model demonstrating the best validation set performance is saved for the test set prediction.
    
    \item \textbf{ChatGPT:} Several recent studies have demonstrated that large language models like ChatGPT can surpass humans in various classification and annotation tasks \cite{10.1001/jamainternmed.2023.2909, Gilardi_2023}. So we explore the scope of ChatGPT \cite{ouyang2022training} to classify ISE in our annotated dataset. To comprehensively assess its capabilities, we explore three distinct settings: zero-shot (ZS), few-shot (FS), and chain-of-thought (CoT) \cite{NEURIPS2022_9d560961} prompting. The chain-of-thought approach gives the model more reasoning about this domain-specific task \cite{min-etal-2022-rethinking}. 
    We thoroughly explored various versions of prompts and refined those that showed encouraging outcomes. We select the optimal prompt through an iterative process of trial and error guided by the empirical observations of the model's output. Our approach involves two prompt templates for conducting the experiments: \textit{`Short'} and \textit{`Long'}. The \textit{`Short'} template offers minimal details concerning the ISE classes, while the \textit{`Long'} variant provides the model with a detailed definition of the classes. We adopt both \textit{`Short'} and \textit{`Long'} templates in the ZS and FS experiments. However, for the chain-of-thought approach, we use a modified version of long prompts, including the reasoning for the examples. We set the temperature value to 0.0 across all experiments to ensure the deterministic behavior of the model. 
    
    The test set \textbf{excludes} all the samples used to identify the best prompts and in-context examples used in the few-shot and chain-of-thought prompts. This ensures unbiased evaluation and prevents the risk of potential data leakage. Due to space constraints, we could not share example prompts in the main paper. However, they are readily accessible through the appendix\footnote{https://tinyurl.com/ycyj3v4a}.
    
\end{itemize}

%\section{Experimental Results}

\section{TREAT-ISE: Benchmark Evaluation}

In this section, we outline the evaluation settings and present the results. We perform comprehensive ablation studies to understand the performance of advanced LLMs, like ChatGPT, on domain-specific, complex text classification task.

\noindent
\textbf{Experimental and Evaluation Setup:} All the experiments were conducted on a GPU-accelerated Google Colab platform. %We used Pandas (1.5.3) and NumPy (1.22.4) to prepare and process the data. 
Machine learning and deep learning models were trained with ktrain \cite{maiya2020ktrain}, while all the transformer models were implemented from Huggingface. Finally, we investigate the performance of the ChatGPT model via API (version \textit{gpt-3.5-turbo-0613}) calls. 

%\footnote{https://huggingface.co/docs/transformers}
TREAT-ISE is partitioned into three mutually exclusive sets: train (80\%), validation (10\%), and test (10\%). We leverage various statistical measures (precision (P), recall (R), F1-score) to asses and understand the model's performance across different classes. The validation set is utilized to tune the model hyperparameters across various experiments. The weighted F1-score (WF1) on the test set is used to compare and determine the superiority of the models.

\subsection{Results}
Table \ref{all-results} presents the classwise performance of all the models on the test set of the TREAT-ISE dataset. BiGRU achieved the highest WF1 of 0.702 among the non-transformer baselines. XLNet outperformed all other models with a maximal WF1 score of 0.774. It excelled particularly well in AM, TP, and RL classes, with scores of 0.83, 0.85, and 0.86, respectively. RoBERTa attained the highest WF1 of 0.757 of the BERT variants. All models encountered challenges in identifying samples from \textit{taking medication} (TM) and \textit{experiencing psychophysical effects} (EP) events. Surprisingly, all the ChatGPT variants underperformed compared to other baselines. We conducted a detailed ablation study to get more insights into this. The CoT prompt acquired the highest WF1 of 0.631 and outperformed all other prompting techniques. In contrast to the other baselines, where there is a balance between classwise precision
and recall, all the ChatGPT variants (except ZS-S) showed much higher recall than precision. This indicates
that ChatGPT tends to overpredict the classes. Overall, the results indicate that identifying treatment information-seeking events is difficult, requires domain knowledge, and has significant room for improvement.

% ChatGPT's over-reliance on prompts/in-context samples and lack of domain knowledge can cause this suboptimal performance. 
% Upon investigating the overall results, it becomes evident that the performance is lackluster compared to other models in similar problem space, which are achieving much higher accuracy \cite{lee-etal-2022-k}. 

\paragraph{Statistical significance:}
We conduct statistical significance testing using the \citeauthor{mcnemar1947note} \shortcite{mcnemar1947note} test to see if the best-performing model (i.e., XLNet) outperforms other models in a statistically meaningful way. Since the results from each classifier are nominal data (i.e., classes of events) and it is difficult to train multiple copies of a model, this test is a suitable approach. We conducted a pair-wise comparison between XLNet and all the other models for each event class. The classwise \textit{P-value} indicates XLNet is significantly better than all other models in three of the five classes, namely, TM (P\textless$0.001$), EP (P=$0.008$), and TP (P=$0.003$). The performance of XLNet is not statistically significant for the remaining two classes. Specifically for AM (P=$0.657$) and RL (P=$0.446$), XLNet's performance is comparable to RoBERTa and FastText, respectively. %The details of the results are added in the appendix.

% for AM (P=$0.657$) and RL (P=$0.446$) for RoBERTa and FastText models. This suggests an insignificant predictive distribution difference between XLNet and RoBERTa in the AM class and between XLNet and FastText in the RL class.
%For the remaining two classes (i.e., AM, RL), the \textit{P-values} are high for AM (P=$0.657$) and RL (P=$0.446$) for RoBERTa and FastText models. This suggests an insignificant predictive distribution difference between XLNet and RoBERTa in the AM class and between XLNet and FastText in the RL class.
%This suggests the predictive distribution difference between XLNet and RoBERTa in the AL class and XLNet and FastText for the RL class is insignificant.

\subsection{Ablation Studies with ChatGPT and XLNet}
\label{ablation}
Although recent studies illustrate that ChatGPT can outperform humans in knowledge-intensive tasks \cite{Gilardi_2023}, results (Table \ref{all-results}) demonstrate that ChatGPT exhibits suboptimal performance in classifying treatment information-seeking events. This is particularly noticeable for samples that require significant domain knowledge to distinguish between events. This motivates us to conduct a deeper investigation into the scope of ChatGPT for such event analysis. We also aim to uncover whether the errors of the transformer models are echoed in ChatGPT or they are distinct. So, in this section, we perform a thorough side-by-side analysis between the best-performing model in our task, XLNet, and the top-performing prompt setting of the ChatGPT model (i.e., chain-of-thought). The findings are as follows.

% Results (Table \ref{all-results}) demonstrate that ChatGPT exhibits suboptimal performance in classifying treatment information-seeking events. However, studies illustrate that ChatGPT can outperform humans in knowledge-intensive tasks \cite{Gilardi_2023}, which motivates us to conduct a deeper investigation into the scope model. We aim to uncover whether the errors of the transformer models echoed in ChatGPT or they are distinct. So, we perform a thorough side-by-side analysis between the best-performing model in our task, XLNet, and the top-performing prompt setting of the ChatGPT model (i.e., chain-of-thought). The findings are as follows.
% We argue that ChatGPT still lacks the capability to handle challenging NLP tasks that require domain knowledge. To get further insights, 
\begin{figure}[h!]
  \centering
  \subfigure{\includegraphics[width=0.32\linewidth]{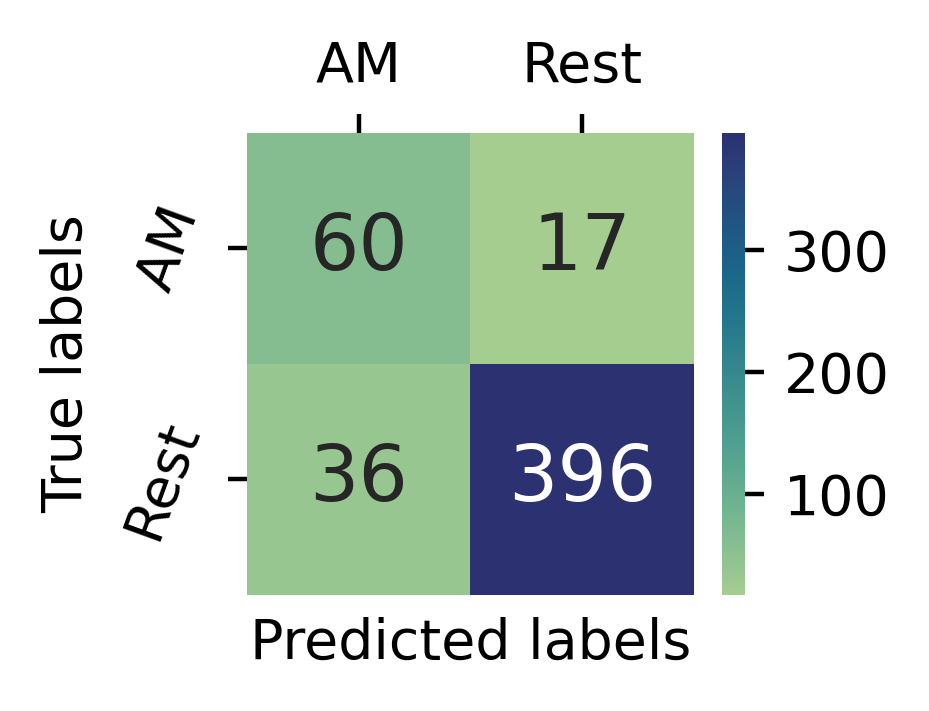}}
  \subfigure{\includegraphics[width=0.32\linewidth]{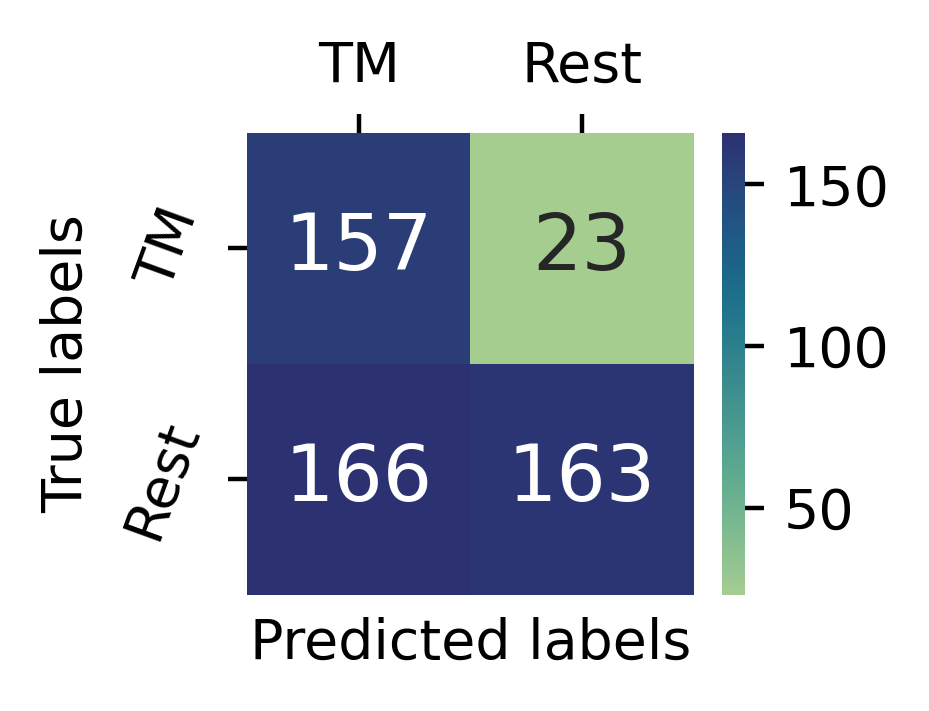}}
    \subfigure{\includegraphics[width=0.32\linewidth]{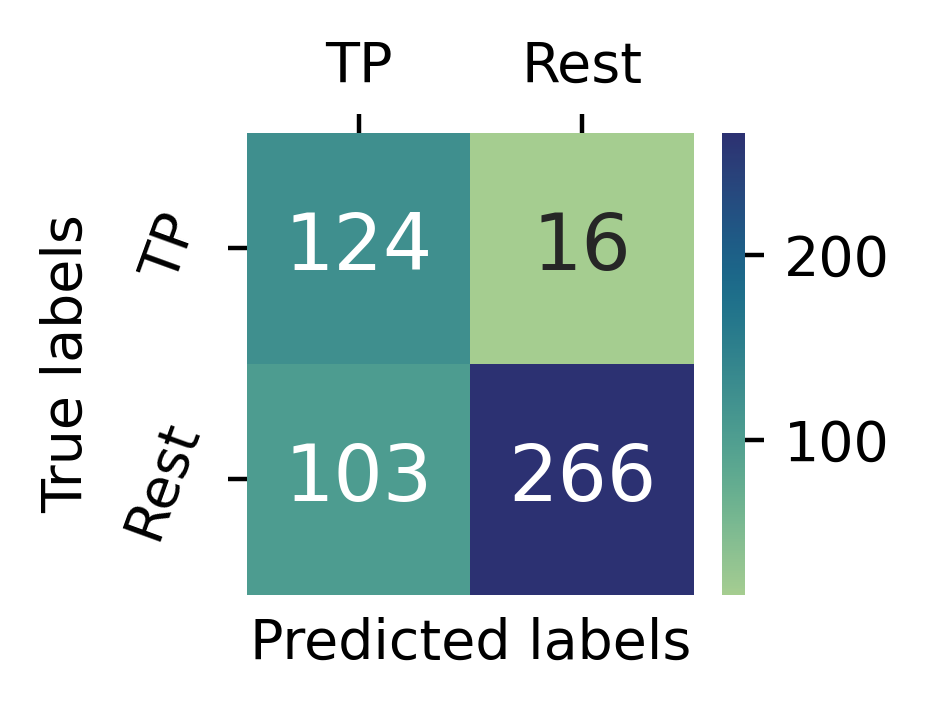}}
  \subfigure{\includegraphics[width=0.32\linewidth]{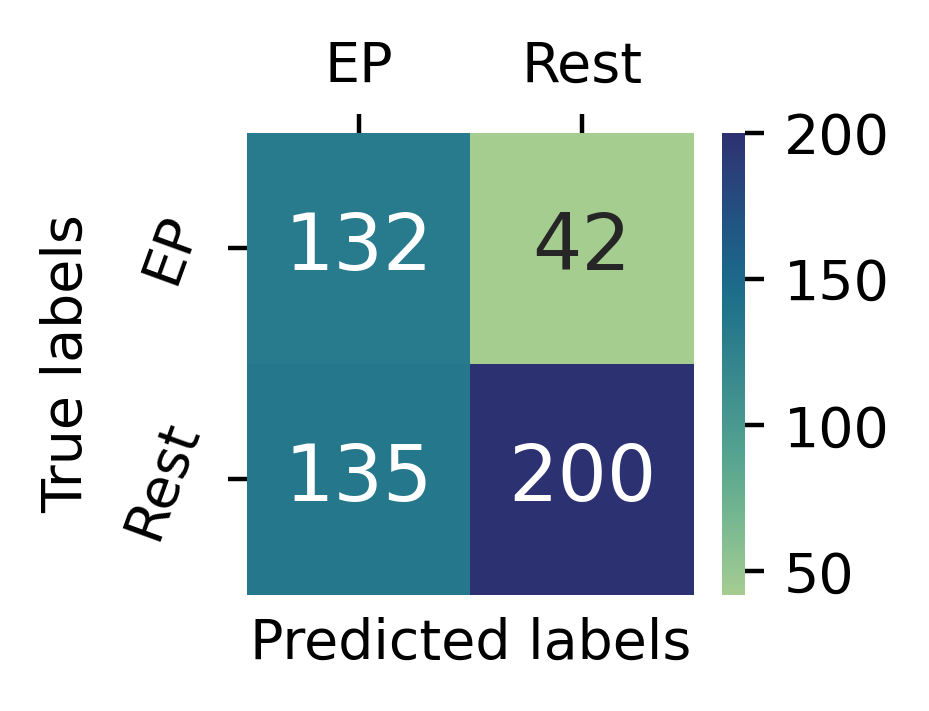}}
  \subfigure{\includegraphics[width=0.32\linewidth]{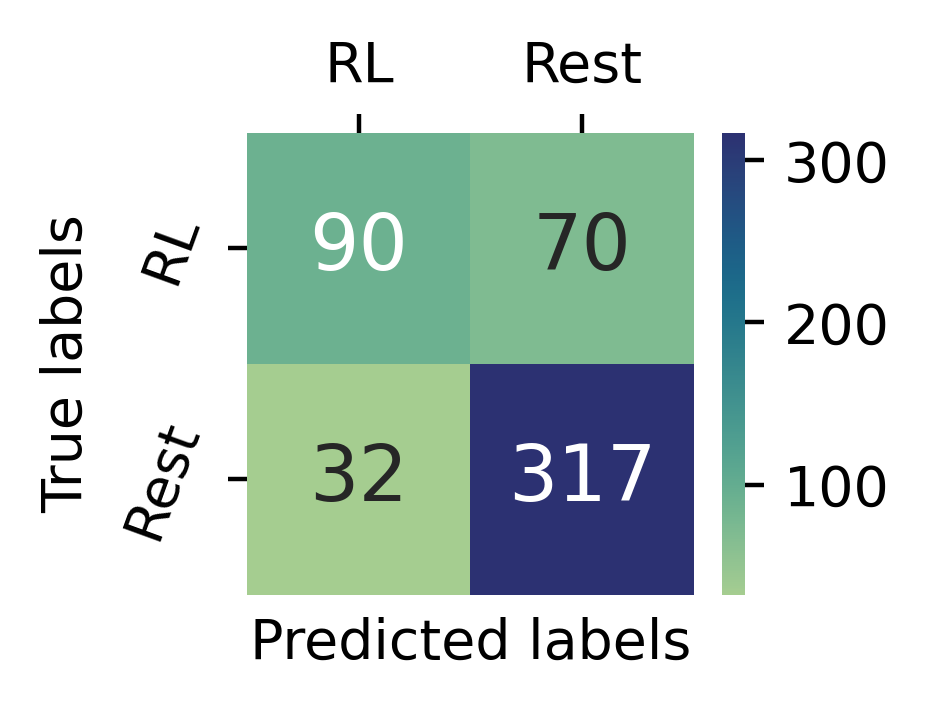}}
  \subfigure{\includegraphics[width=0.32\linewidth]{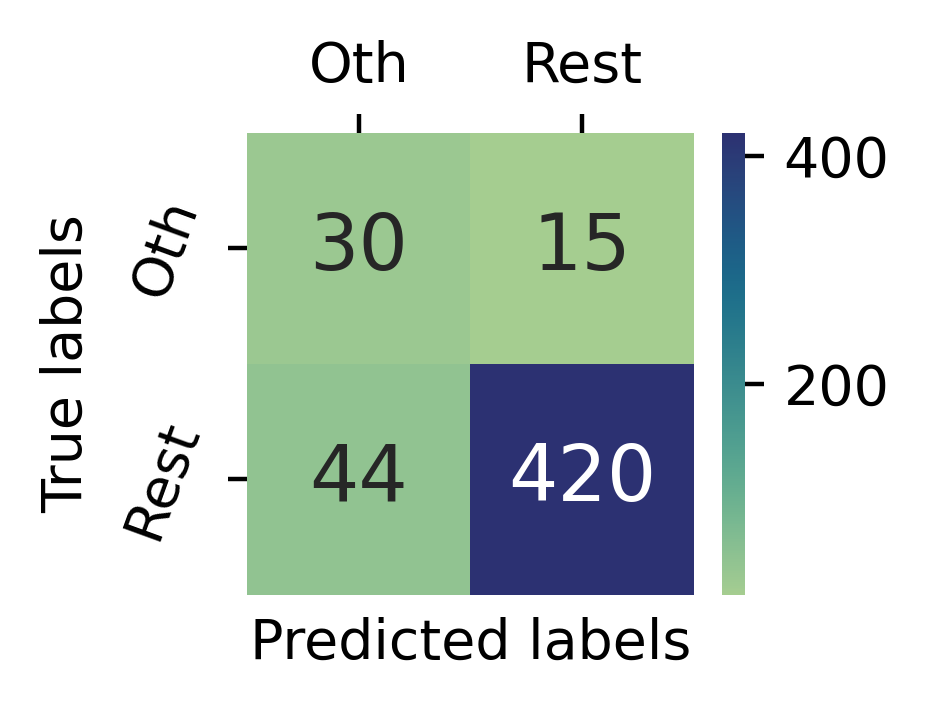}}
 \caption{Confusion matrices of each category for ChatGPT with the chain-of-thought (CoT) approach. The \textit{Rest} class indicates predictions on all other classes.} 
 \label{con-mat}
\end{figure}

\begin{itemize}
\item \textbf{ChatGPT tends to overpredict more:} After qualitative and quantitative analysis, we found that  ChatGPT often fails to understand the context holistically and overpredicts. Consider the following example, which is about \textit{relapse} (RL) and \textit{tapering} (TP). Although ChatGPT predicted these labels correctly because of the mention of side effects and dosage information, it erroneously added TM and EP labels.
    \begin{quote}
        ...I started by quitting kratom completely and taking 2mg of suboxone, I experienced no \textbf{withdrawals} during the switch but also no high. Today I'm down to \textbf{1.5mg of suboxone}, and I'm so happy! Planning to go down to 1.25mg pretty soon too.
    \end{quote}
Figure \ref{con-mat} illustrates the confusion matrices for the ChatGPT chain-of-thought approach. Table \ref{over-pred} presents the classwise overprediction ratio (\#false positive / \#predicted positives) for both ChatGPT and XLNet. Surprisingly, the average overprediction ratio for ChatGPT is 45\%. That means almost half of the time, it incorrectly predicts that samples contain information-seeking events. ChatGPT exhibits higher error in the TM and EP classes, with 166 (out of 323) and 135 (out of 267) mispredictions, respectively. In contrast, XLNet exhibits a drastically lower overprediction ratio for all categories except in the EP class. 
    
\begin{table}[h!]
\centering
\renewcommand*{\arraystretch}{0.92}
\footnotesize
\begin{tabular}{l|cccccc}

  & {\textbf{AM}}& {\textbf{TM}}& {\textbf{TP}} & {\textbf{EP}} & {\textbf{RL}} & \textbf{Oth}\\
 \midrule
\multirow{2}*{CG}&36/96 & 166/323 & 103/227 & 135/267 & 32/122 & 44/74  \\
 & 0.375 & 0.513 & 0.453 & 0.505 & \textbf{0.262} & 0.594  \\
\midrule
 \multirow{2}{*}{XL}&12/75 & 35/165 & 21/139 & 92/227 & 19/155 & 9/33 \\
& 0.160 & 0.212 & 0.151 & 0.405 & \textbf{0.122} & 0.27 \\
\hline

\end{tabular}
\caption{ Classwise overprediction ratio (\#false positive / \#predicted positives) of ChatGPT (CG) with CoT prompts and the XLNet (XL) model. %The ratio is much higher for TM and EP classes $\approx50\%$.
}
\label{over-pred}
\end{table}

% \begin{table}[h!]
% \centering
% \renewcommand*{\arraystretch}{1}
% \footnotesize
% \begin{tabular}{l|C{1.5cm}C{1.5cm}|C{1.5cm}C{1.5cm}}

% & \multicolumn{2}{c}{ChatGPT} & \multicolumn{2}{c}{XLNet}\\
%  \midrule

% \textbf{AM}& 36/96 & 0.375 & 12/75 & 0.160 \\
% \textbf{TM} & 166/323 & 0.513 & 35/165 & 0.212 \\
% \textbf{TP} & 103/227 & 0.453 & 21/139 & 0.151 \\
% \textbf{EP} & 135/267 & 0.505 & 92/227 & 0.405 \\
% \textbf{RL} & 32/122 & 0.262 & 19/155 & 0.122 \\
% \textbf{OTH}& 44/74 & 0.594 & 9/33 & 0.27 \\
% \midrule
% Mean & \multicolumn{2}{c}{0.450} & \multicolumn{2}{c}{0.22}\\
% \hline

% \end{tabular}
% \caption{\label{over-pred} Classwise overprediction ratio (\#false positive / \#predicted positives) of the ChatGPT with CoT approach. The ratio is much higher for TM and EP classes $\approx50\%$.
% }
% \end{table}

\begin{figure}[h!]
  \centering
  \includegraphics[width =0.85\linewidth]{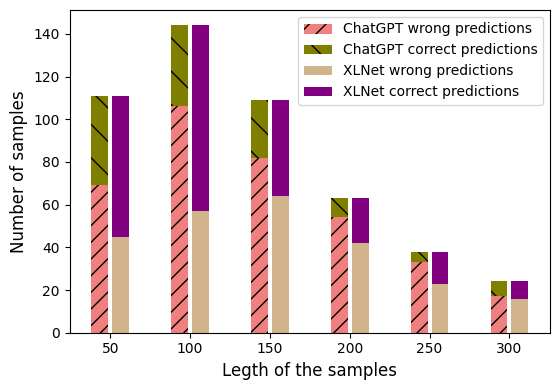}
 \caption{Correlation between sample length and frequency of correct/wrong predictions: as the length of samples (measured in words) increases, the frequencies of accurate predictions decrease for both models.} 
 \label{leng-dis}
\end{figure}

\item \textbf{ChatGPT struggles more on long samples:} For analysis, we compute the frequency of correct and wrong predictions on different length ranges for both ChatGPT (CoT) and XLNet. Figure \ref{leng-dis} illustrates the correlation, indicating that the frequency of accurate prediction is higher among the shorter samples and decreases as sample length increases. On average, the samples where ChatGPT made errors had a length of 128.04, whereas, for XLNet, this value is 140.21. This analysis suggests both models encounter difficulties in understanding information-seeking events with long-range context. However, XLNet shows slightly more robustness than ChatGPT. 

\item \textbf{ChatGPT misclassifies events more:} To obtain the confusion mapping, we calculate the frequency of incorrect predictions for each event in relation to other events as presented in Table \ref{confusion-table}. Analyzing the results, it becomes evident that ChatGPT faces difficulty in distinguishing advice events associated with TM, EP, and RL classes, often misclassifying them as \textit{other} class. The model made the highest (92) number of errors on the RL event class and, most of the time considered it as either the TM or EP event class. Interestingly, XLNet often misclassified TM as EP (10) class.   

\begin{table}[h!]
\renewcommand*{\arraystretch}{0.91}
\centering
\footnotesize
\begin{tabular}
{lc|cccccc|c}
& &{\textbf{AM}}& {\textbf{TM}}& {\textbf{TP}} & {\textbf{EP}} & {\textbf{RL}} & \textbf{Oth} & Total\\
\midrule        
 \multirow{2}{*}{AM} & CG &-  & \textbf{7} & 2 & 3 & 2 & 5 & 19\\
 & XL & - & 1 & 1 & 3 & 2 & 3 & 10 \\
\midrule

 \multirow{2}{*}{TM} &CG & 1 & - & 2 & 2 & 2 & \textbf{10} & 17\\
 & XL& 0 & - & 6 & {10} & 5 & 3 & {24} \\

\midrule
 \multirow{2}{*}{TP} &CG & 2 & 1 & - & 2 & 1 & \textbf{8}  & 14\\
&XL & 0 & 5 & - & 6 & 1 & 0 & 12 \\
\midrule
 \multirow{2}{*}{EP} &CG & 1 & 6 & 5 & - & 3 & \textbf{20} & 35\\
& XL & 1 & 1 & 1 & - & 2 & 0 & 5 \\
\midrule
 \multirow{2}{*}{RL} &CG & 6 & 29 & 15 & \textbf{31} & - & 11 & 92\\
&XL & 0 & 5 & 1 & 6 & - & 1 & 13 \\

\midrule
\multirow{2}{*}{Oth} &CG & 4 & \textbf{10} & 1 & 4 & 2 & - & 21\\
&XL & 7 & 3 & 1 & 1 & 0 & - & 12 \\
\hline

\end{tabular}
\caption{Confusion mapping of ChatGPT (CG) with chain-of-thought approach and XLNet (XL) model. Each cell indicates how many times an event (in row) confuses with another event indicated in the column.   
% \textcolor{red}{[we can discuss do we need this one or if the old table is enough. I have also tried diagonally, which takes more space and does not look good.]}
} 
\label{confusion-table} 
\end{table}

%%Old Table
% \begin{table}[h!]
% %{\arraystretch}{1.2}
% \centering
% \small

% \begin{tabular}
% {l|c|c|c|c|c|c|c}
% \textbf{Class}&AM & TM & TP & EP & RL & OTH & Total\\
% \midrule        
% AM &-  & \underline{7} & 2 & 3 & 2 & 5 & 19\\
% TM & 1 & - & 2 & 2 & 2 & \underline{10} & 17\\
% TP & 2 & 1 & - & 2 & 1 & \underline{8}  & 14\\
% EP & 1 & 6 & 5 & - & 3 & \underline{20} & 35\\
% RL & 6 & 29 & 15 & \underline{31} & - & 11 & \textbf{92}\\
% OTH & 4 & \underline{10}} & 1 & 4 & 2 & - & 21\\
% \hline

% \end{tabular}
% \caption{\label{confusion-table} Confusion mapping of ChatGPT with chain-of-thought  approach. Each cell indicates how many times an event (in row) confuses with another event indicated in the column.   
% }
% \end{table}
The results suggest that ChatGPT is biased toward predicting TM and EP classes. After qualitative observation, we notice that ChatGPT often mislabels samples as TM when dosage information is provided \textit{(e.g., 2mg kratom, 1.5 mg bupe)}, even though these instances do not seek treatment information. Similarly, the model frequently mislabels posts mentioning psychophysical effects (e.g., withdrawals, sleep) as EP, despite these not being information-seeking events. Surprisingly, on 54 occasions, the model identified posts that were seeking treatment information but failed to predict appropriate event classes and mislabeled them as the \textit{other} class. This mislabeling can be attributed to the model's poor understanding of the domain-specific nuances.
\end{itemize}

\section{Conclusion and Future Work}
In this paper, we address a critical social concern by investigating the information needs of individuals who are considering or undergoing recovery from opioid use disorder. On the guidance of experts, we develop a multilabel, multiclass dataset (\textit{TREAT-ISE}) aiming to characterize OUD treatment information-seeking events. This dataset introduces a new resource to the field, enabling us to study MOUD treatment for recovery through the lens of \textit{events}. The event schema we defined can be valuable to surface clinical insights such as knowledge gaps about treatment, tapering strategies, potential misconceptions, and beyond. Moreover, our data collection process, event-centric schema design, and data annotation strategy can be replicated to develop similar resources for other domains. Finally, we benchmark the dataset with a wide range of NLP models and demonstrate the potential challenges of the task with thorough ablation studies. 

There are several scopes for potential improvement. 
Due to costly and time-consuming annotation, we had to limit the dataset size to 5083 samples. We will explore the possibility of minimal supervision to augment the dataset size by leveraging our annotation protocol and additional available data (over 10K samples). Other research can explore how treatment information-seeking events vary in other online communities and subreddits. In addition, investigating how other large models (e.g., GPT-4, LLaMA) perform on this task can provide us with valuable insights. 
% We continue to work towards establishing the practical value of both the dataset and model in effectively understanding opioid use disorder in real-world settings. 

\section{Ethical Considerations}
This research was approved by the Institutional Review Board (IRB) of the author's institution.
\\
\textbf{User Privacy:} All the data samples were collected and annotated in a manner consistent with the terms and conditions of the respective data source. We do not collect or share any personal information (e.g., age, location, gender, identity) that violates the user's privacy. 
\\
\textbf{Biases:} Any biases found in the dataset and model are unintentional. Experts and a set of diverse groups of annotators labeled the data following a comprehensive annotation guideline and all annotations were reviewed to address any potential annotation biases. Our data collection exclusively focused on one subreddit (r/suboxone), possibly leading to a bias towards the r/suboxone community. The developed models can only be used to identify events that we
discussed in the paper. So the chance of using these models
for malicious reasons is very minimal.
% Addressing this data source bias in such social discourse study is challenging \cite{hovy2021five}.
\\
\textbf{Intended Use:} We intend to make our dataset accessible per Reddit policies to encourage further research on online health discourse as well research on MOUD. 
% and social media mining for recovery-related insights. 
% We believe that this dataset and models will be a valuable resource when used in the appropriate manner.
\section{Acknowledgement}
The preparation of this article was partially supported by P30 Center of Excellence grant from the National Institute on Drug Abuse (NIDA) P30DA029926 (PI: Lisa A. Marsch).
% We also thank our data annotators for their rigorous effort. 

\bibliography{aaai24}

\begin{thebibliography}{51}
\providecommand{\natexlab}[1]{#1}

\bibitem[{Balsamo et~al.(2023)Balsamo, Bajardi, De~Francisci~Morales, Monti, and Schifanella}]{Balsamo_Bajardi_De}
Balsamo, D.; Bajardi, P.; De~Francisci~Morales, G.; Monti, C.; and Schifanella, R. 2023.
\newblock The Pursuit of Peer Support for Opioid Use Recovery on Reddit.
\newblock \emph{Proceedings of the International AAAI Conference on Web and Social Media}, 17(1): 12--23.

\bibitem[{Baumgartner et~al.(2020)Baumgartner, Zannettou, Keegan, Squire, and Blackburn}]{baumgartner2020pushshift}
Baumgartner, J.; Zannettou, S.; Keegan, B.; Squire, M.; and Blackburn, J. 2020.
\newblock The Pushshift Reddit Dataset.
\newblock arXiv:2001.08435.

\bibitem[{Bogatinovski et~al.(2022)Bogatinovski, Todorovski, Džeroski, and Kocev}]{BOGATINOVSKI2022117215}
Bogatinovski, J.; Todorovski, L.; Džeroski, S.; and Kocev, D. 2022.
\newblock Comprehensive comparative study of multi-label classification methods.
\newblock \emph{Expert Systems with Applications}, 203: 117215.

\bibitem[{Busetto, Wick, and Gumbinger(2020)}]{busetto2020use}
Busetto, L.; Wick, W.; and Gumbinger, C. 2020.
\newblock How to use and assess qualitative research methods.
\newblock \emph{Neurological Research and practice}, 2: 1--10.

\bibitem[{Chancellor, Baumer, and De~Choudhury(2019)}]{10.1145/3359249}
Chancellor, S.; Baumer, E. P.~S.; and De~Choudhury, M. 2019.
\newblock Who is the "Human" in Human-Centered Machine Learning: The Case of Predicting Mental Health from Social Media.
\newblock \emph{Proc. ACM Hum.-Comput. Interact.}, 3(CSCW).

\bibitem[{Chancellor et~al.(2019{\natexlab{a}})Chancellor, Nitzburg, Hu, Zampieri, and De~Choudhury}]{OpioidRecovery}
Chancellor, S.; Nitzburg, G.; Hu, A.; Zampieri, F.; and De~Choudhury, M. 2019{\natexlab{a}}.
\newblock Discovering Alternative Treatments for Opioid Use Recovery Using Social Media.
\newblock In \emph{Proceedings of the 2019 CHI Conference on Human Factors in Computing Systems}, CHI '19, 1–15. New York, NY, USA: Association for Computing Machinery.
\newblock ISBN 9781450359702.

\bibitem[{Chancellor et~al.(2019{\natexlab{b}})Chancellor, Nitzburg, Hu, Zampieri, and De~Choudhury}]{alternative_chi}
Chancellor, S.; Nitzburg, G.; Hu, A.; Zampieri, F.; and De~Choudhury, M. 2019{\natexlab{b}}.
\newblock Discovering Alternative Treatments for Opioid Use Recovery Using Social Media.
\newblock In \emph{Proceedings of the 2019 CHI Conference on Human Factors in Computing Systems}, CHI '19, 1–15. New York, NY, USA: Association for Computing Machinery.
\newblock ISBN 9781450359702.

\bibitem[{Chen, Johnny, and Conway(2022)}]{CHEN2022100061}
Chen, A.~T.; Johnny, S.; and Conway, M. 2022.
\newblock Examining stigma relating to substance use and contextual factors in social media discussions.
\newblock \emph{Drug and Alcohol Dependence Reports}, 3: 100061.

\bibitem[{Chen, Wang et~al.(2021)}]{chen2021social}
Chen, J.; Wang, Y.; et~al. 2021.
\newblock Social media use for health purposes: systematic review.
\newblock \emph{Journal of medical Internet research}, 23(5): e17917.

\bibitem[{Clark et~al.(2020)Clark, Luong, Le, and Manning}]{clark2020electra}
Clark, K.; Luong, M.-T.; Le, Q.~V.; and Manning, C.~D. 2020.
\newblock ELECTRA: Pre-training Text Encoders as Discriminators Rather Than Generators.
\newblock arXiv:2003.10555.

\bibitem[{Cohen(1960)}]{cohen1960coefficient}
Cohen, J. 1960.
\newblock A coefficient of agreement for nominal scales.
\newblock \emph{Educational and psychological measurement}, 20(1): 37--46.

\bibitem[{Devlin et~al.(2019)Devlin, Chang, Lee, and Toutanova}]{devlin-etal-2019-bert}
Devlin, J.; Chang, M.-W.; Lee, K.; and Toutanova, K. 2019.
\newblock BERT: Pre-training of Deep Bidirectional Transformers for Language Understanding.
\newblock arXiv:1810.04805.

\bibitem[{Dickson-Gomez et~al.(2022)Dickson-Gomez, Spector, Weeks, Galletly, McDonald, and Green~Montaque}]{dickson2022you}
Dickson-Gomez, J.; Spector, A.; Weeks, M.; Galletly, C.; McDonald, M.; and Green~Montaque, H.~D. 2022.
\newblock “You’re not supposed to be on it forever”: medications to treat opioid use disorder (MOUD) related stigma among drug treatment providers and people who use opioids.
\newblock \emph{Substance Abuse: Research and Treatment}, 16: 11782218221103859.

\bibitem[{Edo-Osagie et~al.(2020)Edo-Osagie, {De La Iglesia}, Lake, and Edeghere}]{EDOOSAGIE2020103770}
Edo-Osagie, O.; {De La Iglesia}, B.; Lake, I.; and Edeghere, O. 2020.
\newblock A scoping review of the use of Twitter for public health research.
\newblock \emph{Computers in Biology and Medicine}, 122: 103770.

\bibitem[{Florence, Luo, and Rice(2021)}]{FLORENCE2021108350}
Florence, C.; Luo, F.; and Rice, K. 2021.
\newblock The economic burden of opioid use disorder and fatal opioid overdose in the United States, 2017.
\newblock \emph{Drug and Alcohol Dependence}, 218: 108350.

\bibitem[{Fu et~al.(2023)Fu, Li, Zhou, Li, Lai, Deng, Zhang, Guo, and Wu}]{info:doi/10.2196/43349}
Fu, J.; Li, C.; Zhou, C.; Li, W.; Lai, J.; Deng, S.; Zhang, Y.; Guo, Z.; and Wu, Y. 2023.
\newblock Methods for Analyzing the Contents of Social Media for Health Care: Scoping Review.
\newblock \emph{J Med Internet Res}, 25: e43349.

\bibitem[{Gatto, Basak, and Preum(2023)}]{gatto2023scope}
Gatto, J.; Basak, M.; and Preum, S.~M. 2023.
\newblock Scope of Pre-trained Language Models for Detecting Conflicting Health Information.
\newblock In \emph{Proceedings of the International AAAI Conference on Web and Social Media}, volume~17, 221--232.

\bibitem[{Gilardi, Alizadeh, and Kubli(2023)}]{Gilardi_2023}
Gilardi, F.; Alizadeh, M.; and Kubli, M. 2023.
\newblock {ChatGPT} outperforms crowd workers for text-annotation tasks.
\newblock \emph{Proceedings of the National Academy of Sciences}, 120(30).

\bibitem[{Joulin et~al.(2016)Joulin, Grave, Bojanowski, and Mikolov}]{joulin-etal-2017-bag}
Joulin, A.; Grave, E.; Bojanowski, P.; and Mikolov, T. 2016.
\newblock Bag of Tricks for Efficient Text Classification.
\newblock arXiv:1607.01759.

\bibitem[{Kanchan and Gaidhane(2023)}]{kanchan2023social}
Kanchan, S.; and Gaidhane, A. 2023.
\newblock Social Media Role and Its Impact on Public Health: A Narrative Review.
\newblock \emph{Cureus}, 15(1).

\bibitem[{Lavertu, Hamamsy, and Altman(2021)}]{Lavertu2021.04.01.21254815}
Lavertu, A.; Hamamsy, T.; and Altman, R.~B. 2021.
\newblock Monitoring the opioid epidemic via social media discussions.
\newblock \emph{medRxiv}.

\bibitem[{Li et~al.(2023)Li, Fang, Yang, Wang, Ye, Zhao, and Zhang}]{li2023evaluating}
Li, B.; Fang, G.; Yang, Y.; Wang, Q.; Ye, W.; Zhao, W.; and Zhang, S. 2023.
\newblock Evaluating ChatGPT's Information Extraction Capabilities: An Assessment of Performance, Explainability, Calibration, and Faithfulness.
\newblock arXiv:2304.11633.

\bibitem[{Li, Ji, and Han(2021)}]{li-etal-2021-document}
Li, S.; Ji, H.; and Han, J. 2021.
\newblock Document-Level Event Argument Extraction by Conditional Generation.
\newblock arXiv:2104.05919.

\bibitem[{Liu et~al.(2019{\natexlab{a}})Liu, Li, Zhang, Yang, and Zhou}]{liu-etal-2019-event}
Liu, S.; Li, Y.; Zhang, F.; Yang, T.; and Zhou, X. 2019{\natexlab{a}}.
\newblock Event Detection without Triggers.
\newblock In \emph{Proceedings of the 2019 Conference of the North {A}merican Chapter of the Association for Computational Linguistics: Human Language Technologies, Volume 1 (Long and Short Papers)}, 735--744. Minneapolis, Minnesota.

\bibitem[{Liu et~al.(2019{\natexlab{b}})Liu, Ott, Goyal, Du, Joshi, Chen, Levy, Lewis, Zettlemoyer, and Stoyanov}]{liu2019roberta}
Liu, Y.; Ott, M.; Goyal, N.; Du, J.; Joshi, M.; Chen, D.; Levy, O.; Lewis, M.; Zettlemoyer, L.; and Stoyanov, V. 2019{\natexlab{b}}.
\newblock RoBERTa: A Robustly Optimized BERT Pretraining Approach.
\newblock arXiv:1907.11692.

\bibitem[{Lu et~al.(2021)Lu, Lin, Xu, Han, Tang, Li, Sun, Liao, and Chen}]{lu-etal-2021-text2event}
Lu, Y.; Lin, H.; Xu, J.; Han, X.; Tang, J.; Li, A.; Sun, L.; Liao, M.; and Chen, S. 2021.
\newblock {T}ext2{E}vent: Controllable Sequence-to-Structure Generation for End-to-end Event Extraction.
\newblock In \emph{Proceedings of the 59th Annual Meeting of the Association for Computational Linguistics and the 11th International Joint Conference on Natural Language Processing (Volume 1: Long Papers)}, 2795--2806. Online.

\bibitem[{Ma et~al.(2023)Ma, Taylor, Wang, and Peng}]{ma-etal-2023-dice}
Ma, M.~D.; Taylor, A.~K.; Wang, W.; and Peng, N. 2023.
\newblock DICE: Data-Efficient Clinical Event Extraction with Generative Models.
\newblock arXiv:2208.07989.

\bibitem[{Maiya(2020)}]{maiya2020ktrain}
Maiya, A.~S. 2020.
\newblock ktrain: A Low-Code Library for Augmented Machine Learning.
\newblock \emph{arXiv preprint arXiv:2004.10703}.

\bibitem[{McNemar(1947)}]{mcnemar1947note}
McNemar, Q. 1947.
\newblock Note on the sampling error of the difference between correlated proportions or percentages.
\newblock \emph{Psychometrika}, 12(2): 153--157.

\bibitem[{Min et~al.(2022)Min, Lyu, Holtzman, Artetxe, Lewis, Hajishirzi, and Zettlemoyer}]{min-etal-2022-rethinking}
Min, S.; Lyu, X.; Holtzman, A.; Artetxe, M.; Lewis, M.; Hajishirzi, H.; and Zettlemoyer, L. 2022.
\newblock Rethinking the Role of Demonstrations: What Makes In-Context Learning Work?
\newblock In \emph{Proceedings of the 2022 Conference on Empirical Methods in Natural Language Processing}, 11048--11064. Abu Dhabi, United Arab Emirates.

\bibitem[{Mirzaei, Meshgi, and Sekine(2023)}]{mirzaei-etal-2023-real}
Mirzaei, M.~S.; Meshgi, K.; and Sekine, S. 2023.
\newblock What is the Real Intention behind this Question? Dataset Collection and Intention Classification.
\newblock In \emph{Proceedings of the 61st Annual Meeting of the Association for Computational Linguistics (Volume 1: Long Papers)}, 13606--13622. Toronto, Canada: Association for Computational Linguistics.

\bibitem[{Mooney et~al.(2020)Mooney, Valdez, Cousins, Yoo, Zhu, and Hser}]{mooney2020patient}
Mooney, L.~J.; Valdez, J.; Cousins, S.~J.; Yoo, C.; Zhu, Y.; and Hser, Y.-I. 2020.
\newblock Patient decision aid for medication treatment for opioid use disorder (PtDA-MOUD): Rationale, methodology, and preliminary results.
\newblock \emph{Journal of Substance Abuse Treatment}, 108: 115--122.

\bibitem[{Naik, Bogart, and Rose(2017)}]{naik-etal-2017-extracting}
Naik, A.; Bogart, C.; and Rose, C. 2017.
\newblock Extracting Personal Medical Events for User Timeline Construction using Minimal Supervision.
\newblock In \emph{{B}io{NLP} 2017}, 356--364. Vancouver, Canada,.

\bibitem[{Neely, Eldredge, and Sanders(2021)}]{Neely2021-ji}
Neely, S.; Eldredge, C.; and Sanders, R. 2021.
\newblock Health information seeking behaviors on social media during the {COVID-19} pandemic among American social networking site users: Survey study.
\newblock \emph{J. Med. Internet Res.}, 23(6): e29802.

\bibitem[{Olsen and Sharfstein(2014)}]{10.1001/jama.2014.2147}
Olsen, Y.; and Sharfstein, J.~M. 2014.
\newblock {Confronting the Stigma of Opioid Use Disorder—and Its Treatment}.
\newblock \emph{JAMA}, 311(14): 1393--1394.

\bibitem[{Ouyang et~al.(2022)Ouyang, Wu, Jiang, Almeida, Wainwright, Mishkin, Zhang, Agarwal, Slama, Ray, Schulman, Hilton, Kelton, Miller, Simens, Askell, Welinder, Christiano, Leike, and Lowe}]{ouyang2022training}
Ouyang, L.; Wu, J.; Jiang, X.; Almeida, D.; Wainwright, C.~L.; Mishkin, P.; Zhang, C.; Agarwal, S.; Slama, K.; Ray, A.; Schulman, J.; Hilton, J.; Kelton, F.; Miller, L.; Simens, M.; Askell, A.; Welinder, P.; Christiano, P.; Leike, J.; and Lowe, R. 2022.
\newblock Training language models to follow instructions with human feedback.
\newblock arXiv:2203.02155.

\bibitem[{Patwa et~al.(2021)Patwa, Sharma, Pykl, Guptha, Kumari, Akhtar, Ekbal, Das, and Chakraborty}]{Patwa_2021}
Patwa, P.; Sharma, S.; Pykl, S.; Guptha, V.; Kumari, G.; Akhtar, M.~S.; Ekbal, A.; Das, A.; and Chakraborty, T. 2021.
\newblock Fighting an Infodemic: {COVID}-19 Fake News Dataset.
\newblock In \emph{Combating Online Hostile Posts in Regional Languages during Emergency Situation}, 21--29.

\bibitem[{Romano et~al.(2024)Romano, Sharif, Basak, Gatto, and Preum}]{romano2023themedriven}
Romano, W.; Sharif, O.; Basak, M.; Gatto, J.; and Preum, S. 2024.
\newblock Theme-driven Keyphrase Extraction to Analyze Social Media Discourse.
\newblock \emph{Proceedings of the International AAAI Conference on Web and Social Media}.

\bibitem[{Sanh et~al.(2020)Sanh, Debut, Chaumond, and Wolf}]{sanh2020distilbert}
Sanh, V.; Debut, L.; Chaumond, J.; and Wolf, T. 2020.
\newblock DistilBERT, a distilled version of BERT: smaller, faster, cheaper and lighter.
\newblock arXiv:1910.01108.

\bibitem[{Sharma et~al.(2020)Sharma, Choudhury, Althoff, and Sharma}]{sharma2020engagement}
Sharma, A.; Choudhury, M.; Althoff, T.; and Sharma, A. 2020.
\newblock Engagement Patterns of Peer-to-Peer Interactions on Mental Health Platforms.
\newblock arXiv:2004.04999.

\bibitem[{Skaik and Inkpen(2020)}]{acm_review}
Skaik, R.; and Inkpen, D. 2020.
\newblock Using Social Media for Mental Health Surveillance: A Review.
\newblock \emph{ACM Comput. Surv.}, 53(6).

\bibitem[{Song et~al.(2020)Song, Tan, Qin, Lu, and Liu}]{song2020mpnet}
Song, K.; Tan, X.; Qin, T.; Lu, J.; and Liu, T.-Y. 2020.
\newblock MPNet: Masked and Permuted Pre-training for Language Understanding.
\newblock arXiv:2004.09297.

\bibitem[{Strong et~al.(2023)Strong, DiGiammarino, Weng, Kumar, Hosamani, Hom, and Chen}]{10.1001/jamainternmed.2023.2909}
Strong, E.; DiGiammarino, A.; Weng, Y.; Kumar, A.; Hosamani, P.; Hom, J.; and Chen, J.~H. 2023.
\newblock {Chatbot vs Medical Student Performance on Free-Response Clinical Reasoning Examinations}.
\newblock \emph{JAMA Internal Medicine}.

\bibitem[{Tibshirani and Manning(2014)}]{tibshirani-manning-2014-robust}
Tibshirani, J.; and Manning, C.~D. 2014.
\newblock Robust Logistic Regression using Shift Parameters.
\newblock In \emph{Proceedings of the 52nd Annual Meeting of the Association for Computational Linguistics (Volume 2: Short Papers)}, 124--129. Baltimore, Maryland.

\bibitem[{van Aken et~al.(2018)van Aken, Risch, Krestel, and Löser}]{van-aken-etal-2018-challenges}
van Aken, B.; Risch, J.; Krestel, R.; and Löser, A. 2018.
\newblock Challenges for Toxic Comment Classification: An In-Depth Error Analysis.
\newblock arXiv:1809.07572.

\bibitem[{Vaswani et~al.(2017)Vaswani, Shazeer, Parmar, Uszkoreit, Jones, Gomez, Kaiser, and Polosukhin}]{NIPS2017_3f5ee243}
Vaswani, A.; Shazeer, N.; Parmar, N.; Uszkoreit, J.; Jones, L.; Gomez, A.~N.; Kaiser, L.~u.; and Polosukhin, I. 2017.
\newblock Attention is All you Need.
\newblock In Guyon, I.; Luxburg, U.~V.; Bengio, S.; Wallach, H.; Fergus, R.; Vishwanathan, S.; and Garnett, R., eds., \emph{Advances in Neural Information Processing Systems}, volume~30.

\bibitem[{Wang and Manning(2012)}]{wang-manning-2012-baselines}
Wang, S.; and Manning, C. 2012.
\newblock Baselines and Bigrams: Simple, Good Sentiment and Topic Classification.
\newblock In \emph{Proceedings of the 50th Annual Meeting of the Association for Computational Linguistics (Volume 2: Short Papers)}, 90--94.

\bibitem[{Wei et~al.(2022)Wei, Wang, Schuurmans, Bosma, ichter, Xia, Chi, Le, and Zhou}]{NEURIPS2022_9d560961}
Wei, J.; Wang, X.; Schuurmans, D.; Bosma, M.; ichter, b.; Xia, F.; Chi, E.; Le, Q.~V.; and Zhou, D. 2022.
\newblock Chain-of-Thought Prompting Elicits Reasoning in Large Language Models.
\newblock In Koyejo, S.; Mohamed, S.; Agarwal, A.; Belgrave, D.; Cho, K.; and Oh, A., eds., \emph{Advances in Neural Information Processing Systems}, volume~35, 24824--24837.

\bibitem[{Wen and Rose(2012)}]{wen_12}
Wen, M.; and Rose, C.~P. 2012.
\newblock Understanding Participant Behavior Trajectories in Online Health Support Groups Using Automatic Extraction Methods.
\newblock In \emph{Proceedings of the 2012 ACM International Conference on Supporting Group Work}, GROUP '12, 179–188. New York, NY, USA: Association for Computing Machinery.
\newblock ISBN 9781450314862.

\bibitem[{Wen et~al.(2013)Wen, Zheng, Jang, Xiang, and Penstein~Ros{\'e}}]{wen-etal-2013-extracting}
Wen, M.; Zheng, Z.; Jang, H.; Xiang, G.; and Penstein~Ros{\'e}, C. 2013.
\newblock Extracting Events with Informal Temporal References in Personal Histories in Online Communities.
\newblock In \emph{Proceedings of the 51st Annual Meeting of the Association for Computational Linguistics (Volume 2: Short Papers)}, 836--842. Sofia, Bulgaria.

\bibitem[{Yang et~al.(2020)Yang, Dai, Yang, Carbonell, Salakhutdinov, and Le}]{yang2020xlnet}
Yang, Z.; Dai, Z.; Yang, Y.; Carbonell, J.; Salakhutdinov, R.; and Le, Q.~V. 2020.
\newblock XLNet: Generalized Autoregressive Pretraining for Language Understanding.
\newblock arXiv:1906.08237.

\end{thebibliography}

\end{document}